\documentclass[10pt, a4paper]{article}
\usepackage{lrec}
\usepackage{multibib}
\newcites{languageresource}{Language Resources}
\usepackage{graphicx}
\usepackage{tabularx}
\usepackage{soul}
\usepackage{caption}

\usepackage{epstopdf}
\usepackage[utf8]{inputenc}

\usepackage{hyperref}
\usepackage{xstring}

\usepackage{color}
\usepackage[lf]{berenis}

\title{SpokesBiz -- an Open Corpus of Conversational Polish \\}

\name{\\\textbf{Piotr P\c{e}zik\textsuperscript{1}, Sylwia Karasińska\textsuperscript{1}, Anna Cichosz\textsuperscript{1},}\\ 
\textbf{Łukasz Jałowiecki\textsuperscript{1}, Konrad Kaczyński\textsuperscript{1}, Małgorzata Krawentek\textsuperscript{1},} \\
\textbf{
Karolina Walkusz\textsuperscript{1}, Paweł Wilk\textsuperscript{1}, Mariusz Kleć\textsuperscript{2}, Krzysztof Szklanny\textsuperscript{2}, Szymon Marszałkowski\textsuperscript{3}}} 

\address{\textsuperscript{1}University of Łódź, \textsuperscript{2}Polish-Japanese Academy of Information Technology, \textsuperscript{2}Voicelab.AI\\
piotr.pezik@uni.lodz.pl}

\abstract{
This paper announces the early release of SpokesBiz, a freely available corpus of conversational Polish developed within the CLARIN-BIZ project and comprising over 650 hours of recordings. The transcribed recordings have been diarized and manually annotated for punctuation and casing. We outline the general structure and content of the corpus, showcasing selected applications in linguistic research, evaluation and improvement of automatic speech recognition (ASR) systems. 
\\ \newline  }

\begin{document}

\maketitleabstract

\section{Motivation}

Despite recent advances in Automatic Speech Recognition (ASR) and Large Language Models (LLMs) corpora of spoken language continue to play an important role in the development and evaluation of language processing and generation tools. For one thing, the performance of ASR systems can only be independently evaluated on new corpora which eliminate the possibility of test data leaks and result manipulation. Additionally, the quality of state-of-the-art speech recognition is simply insufficient in areas of spoken language research where correct transcriptions of unaccented function words or recently popularized named entities are important.
This paper reports the early release of SpokesBiz, a freely available corpus of conversational Polish developed within the CLARIN-BIZ project which increases the accessibility of Polish-language speech resources for the areas and applications briefly signalled above.  

\section{SpokesBiz}

The Spokes corpora are a family of Polish speech datasets which have been growing steadily since the early 2000s. Historically, the first recordings included in the 2015 version of Spokes (SpokesPL) \cite{pezik_spokes_2015} had originally been collected in the PELCRA Reference Corpus \cite{walinski_web_2007} and the National Corpus of Polish projects
\cite{pezik_jezyk_2012}. A number of subsequent extensions were added  in the CLARIN-PL project \cite{pezik_increasing_2018}\footnote{See \url{http://docs.pelcra.pl/doku.php?id=spoken_offline_corpora} for a downloadable subset of SpokesPL.}. The first spoken corpus released in the more recent CLARIN-BIZ project was DiaBiz\footnote{See also \url{http://docs.pelcra.pl/doku.php?id=diabiz}} \cite{pezik_diabiz_2022} is a commercially offered dataset of approximately 4000 loosely scripted conversations representing a variety of customer interactions in 9 business domains\footnote{See \url{ http://docs.pelcra.pl/doku.php?id=diabiz}}. SpokesBiz is a separate, openly available corpus of spoken Polish developed in the CLARIN-BIZ project at the University of Lodz in cooperation with Wroclaw University of Science and Technology. 

\section{Data Acquisition and Structure}

The seven main subsets of SpokesBiz differ in terms of the data acquisition scenarios: 

\begin{itemize}
  \item  CBIZ\_BIO is a set of face-to-face biographical interviews, informal in character, following a past-present-future scenario (the interviewees’ childhood and school experiences, current job and family situation and plans for the future);
  
    \item CBIZ\_INT: job interviews with potential babysitters;

    \item CBIZ\_LUZ: free and natural conversations among friends and families, no topic restrictions;
    
    \item CBIZ\_POD: a collection of Internet podcasts on board games\footnote{\url{https://gradanie.znadplanszy.pl}}, nature photography \footnote{\url{https://www.podkasty.info/katalog/podkast/3699-Spotkania_z_Przyroda}}, society and travelling \footnote{\url{https://www.youtube.com/c/BrzmienieSwiatazlotuDrozda}}  and international affairs\footnote{\url{https://dzialzagraniczny.pl}};

    \item CBIZ\_PRES: student presentations concerning a variety of academic and popular topics (culture, literature, parenting, gender roles) 

    \item CBIZ\_VC \& CBIZ\_VC2: thematic discussions on various topics related to society and lifestyle (e.g. coronavirus, legalising drugs, Brexit, limits of democracy, losing weight, advantages of living in the city and in the country); 

    \item CBIZ\_WYW: interviews based on a set list of questions (e.g. ways of spending free time, pets, holidays, favourite books/films, first job, children, visits abroad).
\end{itemize}

Table \ref{Tab:sbiz_structure} shows the current size of each of the subcorpora specified above in terms of the number of speakers, utterances and length of recordings. 

\begin{table*}[ht]
\begin{center}

\begin{tabular}{|l|l|l|l|l|l|}
   
      \hline
      Subcorpus & Recordings & Words & Utterances & Hours & Speakers \\
      \hline\hline
CBIZ\_VC & 168	& 1 416 210	& 91 585 & 160 & 196 \\ 
CBIZ\_BIO & 170 &	1 383 646 &	68 429 & 166 & 170 \\ 
CBIZ\_LUZ & 297	& 1 510 024	& 103 571 & 157 & 116 \\ 
CBIZ\_POD &	178 & 991 464 & 47 221 & 92 & 12 \\ 
CBIZ\_WYW & 46	& 327 148	& 16 778 & 37 & 46 \\ 
CBIZ\_INT & 10 &	26 006	& 1 575 & 2 & 11 \\ 
CBIZ\_PRES & 56 &	256 922 & 18 120 & 38 & 39 \\ 

      \hline
Total & 925 & 5 911 420 & 347 279 & 652 & 590 \\
      \hline

\end{tabular}
\caption{The structure of SpokesBiz. The last cell in the bottom row shows the number of distinct speakers across all the subcorpora.}
 \label{Tab:sbiz_structure}
\end{center}
\end{table*}

\begin{table*}[h!]
\begin{center}
\begin{tabular}{|l|l|l|l|}
   
      \hline
      Gender & No. of speakers  & No. of words & Proportions\\
      \hline\hline
female & 348  &	3 163 350  & 53.51\%\\ 
male & 204  &	2 748 070 & 46.49\%\\ 
not declared & 0	& 0  & ---\\ 

      \hline
Total & 552 & 5 911 420  & 100\%\\
      \hline

\end{tabular}
\caption{Gender of SpokesBiz speakers }
\label{Tab:sbiz_gender}
\end{center}
\end{table*}

\begin{table*}[h!]
\begin{center}
\begin{tabular}{|l|l|l|l|l|}
   
      \hline
      Age & No. of speakers  & No. of words & Proportions\\
      \hline\hline
below 20  & 42   &	235 357  & 3.98\% \\ 
20-29 & 211   &	1 745 886  & 29.53\%\\ 
30-39 & 146  & 2 239 053  & 37.88\%  \\ 
40-49 & 64  & 1 092 547  & 18.48\%  \\ 
50-59 &  30  &  190 290  & 3.22\%   \\ 
60-69 &  34  &  219 580  & 3.72\%   \\ 
70 and more  &  25  &  188 707  & 3.19\%   \\ 

      \hline
Total & 552 & 5 911 420 & 100\% \\
      \hline

\end{tabular}
\caption{Age of SpokesBiz speakers }
\label{Tab:sbiz_age}
\end{center}
\end{table*}

\begin{table*}[h!]
\begin{center}
\begin{tabular}{|l|l|l|l|}
   
      \hline
      Education & No. of speakers  & No. of words & Proportions\\
      \hline\hline
primary  & 35   &	187 901  & 3.18\%  \\ 
secondary & 182     &	1 288 127  & 21.79\% \\ 
higher & 323  & 4 373 333  & 73.98\%  \\ 
vocational & 11	& 62 034  & 1.05\%   \\ 
none &  1	& 25  & 0.00\%   \\ 
      \hline
Total & 552 & 5 911 420  & 100\%\\
      \hline

\end{tabular}
\caption{Education of SpokesBiz speakers }
\label{Tab:sbiz_edu}
\end{center}
\end{table*}

Table \ref{Tab:sbiz_gender} shows the declared gender of speakers in SpokesBiz. As can be seen, the data are well-balanced between male and female language users, with a slight majority uttered by the latter (54\%). 

As far as age is concerned, Table \ref{Tab:sbiz_age} shows the proportions between different age groups of speakers in SpokesBiz. For formal reasons (including regulations concerning parental consent), the great majority of SpokesBiz speakers are over 18. Most of them are in their 20s or 30s (and their output covers 33.60\% and 32.99\% of the corpus respectively) but older age groups are also represented.

Another variable taken into account in the metadata is the level of education. Table \ref{Tab:sbiz_edu} shows that most SpokesBiz speakers have received higher or secondary education (which covers 71.13\% and 23.99\% of the SpokesBiz word count respectively); speakers with primary and vocational education are present but considerably less represented present in the corpus.

\section{Annotation and metadata}

The original recordings were automatically diarized and transcribed using a combination of ASR systems (Voicelab\footnote{https://voicelab.ai/asr-api-web-socket-grpc-http} and Whisper\footnote{https://github.com/linto-ai/whisper-timestamped} \cite{radford2022robust,JSSv031i07} paired with the pyannote diarization tool \footnote{https://github.com/pyannote/pyannote-audio}) and loaded into a dedicated manual transcription platform. The automatic transcripts were manually corrected and punctuated \cite{9687976} using guidelines developed for the DiaBiz corpus.
After manual corrections, the transcriptions were re-aligned automatically using Voicelab's ASR engine (a.k.a. Moonlight) at the level of utterances and words and phonemes. At the time of writing this paper, only the utterance and word alignments are made available. 
The diarized transcriptions are also annotated with basic information about the speakers such as the age range, gender and place of residence. In other words, leaving aside occasional misalignements, each word in the corpus can be associated with a particular speaker. This in turn opens the possibility of running sociolectal analyses of the corpus (see Section \ref{section:corp_based_analyses}).  








\section{Availability}

SpokesBiz is currently released free of charge for non-commercial purposes. The instructions for accessing SpokesBiz are available at \url{http://docs.pelcra.pl/doku.php?id=spokesbiz}. In short, to access the transcriptions and corresponding recordings, one has to fill in a simple form.

\section{Use cases}
\subsection{ASR evaluation}

One of the envisaged use-cases of SpokesBiz is the evaluation and fine-tuning of ASR systems. To illustrate this application of the corpus, we present the results of evaluating the Whisper large-v2 ASR model on different subsets of SpokesBiz. We report a number of ASR metrics which measure the edit distance between the automatic and manual transcriptions including: WER (word error rate), i.e. the proportion of all word errors to words processed, MER (match error rate), which is defined as the proportion of words that were incorrectly predicted and inserted, and WIL (word information lost) indicating the percentage of characters that were incorrectly predicted. The results are shown in Table \ref{Tab:sbiz_wer_eval}. It is interesting to note that student presentations (represented by the CBIZ\_PRES subcorpus of SpokesBiz) involving only one speaker speaking at a regular pace posed a much smaller challenge for the ASR engine (15.2\% WER) than dynamic multi-speaker podcasts containing niche video-game related vocabulary (CBIZ\_POD - 26\% WER). This result lends weight to the more general observation that officially reported WER rates for ASR solutions should be taken with a grain of salt as they may vary significantly for different types of spoken language.

\begin{table*}[ht]
\begin{center}

\begin{tabular}{|l|l|l|l|l|l|}

    \hline
    Subcorpus & WER & MER & WIL & WER\_stdev & No. of recordings \\
    \hline\hline
CBIZ\_BIO     & 0.208 & 0.202 & 0.202 & 0.074 & 10 \\
CBIZ\_LUZ     & 0.182 & 0.18  & 0.18  & 0.04  & 10 \\
CBIZ\_PRES    & 0.152 & 0.152 & 0.152 & 0.055 & 10 \\
CBIZ\_VC      & 0.201 & 0.198 & 0.198 & 0.041 & 20 \\
CBIZ\_PODCAST & 0.26  & 0.252 & 0.252 & 0.065 & 10 \\
CBIZ\_INT     & 0.192 & 0.192 & 0.192 & 0.024 & 9  \\
CBIZ\_WYW     & 0.21  & 0.209 & 0.209 & 0.036 & 10 \\ 

    \hline
All samples & 0.201 & 0.198 & 0.198 & 0.056 & 79 \\
    \hline

\end{tabular}
\caption{ASR accuracy for recordings sampled from SpokesBiz.}
 \label{Tab:sbiz_wer_eval}
\end{center}
\end{table*}

\begin{figure}[!h]
\begin{center}
\includegraphics[scale=0.5]{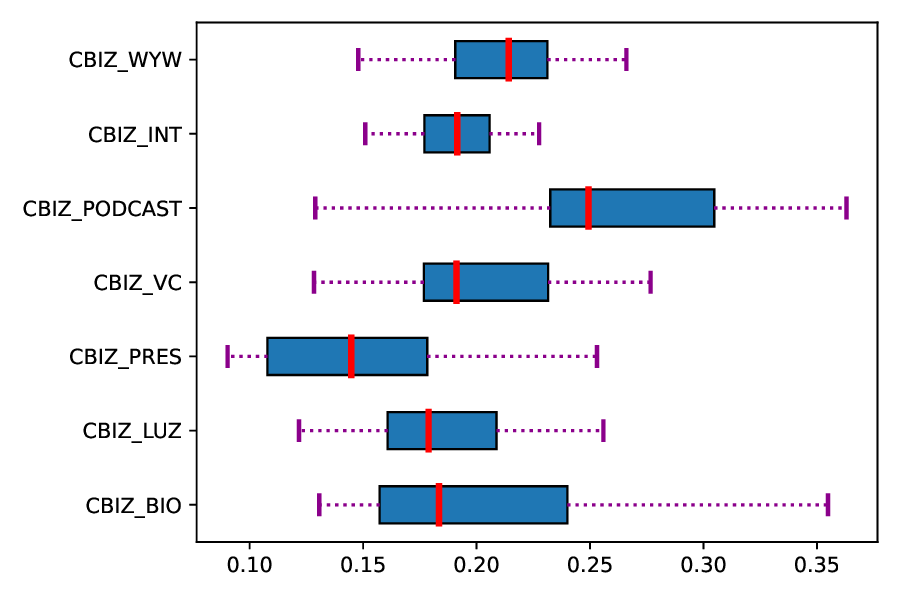} 
\caption{WER distribution for subcorpora samples.}
\label{Fig:fig.1}
\end{center}
\end{figure}


\subsection{Corpus-based analyses}
\label{section:corp_based_analyses}

\begin{figure*}
    \centering

    \includegraphics[scale=0.85]{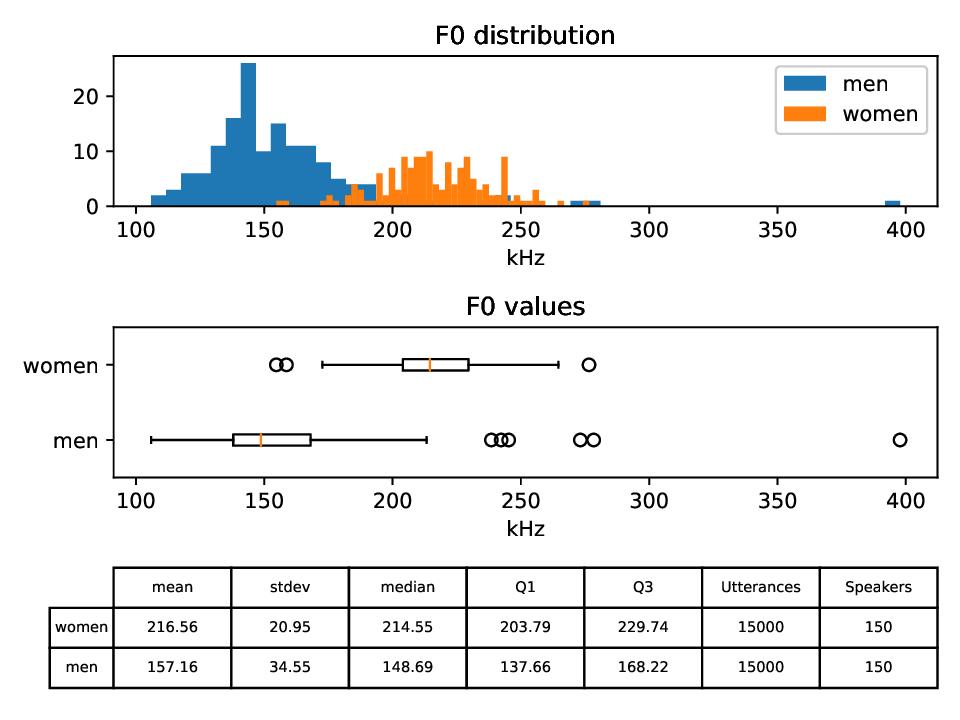}
    \caption{Gender-based differences in pitch values in SpokesBiz}
    \label{corpus_sex}
\end{figure*}

The availability of utterance-level speaker metadata in SpokesBiz makes it possible to run various corpus linguistic analyses on this dataset. For example, it is possible to utilize the corpus metadata and time-alignment annotation  to compare gender-related differences in the average fundamental frequency of male vs female voices. Using 100 utterance samples from 150 male \& 150 female speakers we created a balanced subcorpus of 30k utterance segments. We then computed the fundamental frequency (F0) for each segment with Parselmouth\footnote{https://parselmouth.readthedocs.io, try it out \href{https://colab.research.google.com/drive/1mUplcFnhaBEOLJkHltf1jEr45eSQBPXj?usp=sharing} {here}} and averaged the values across the speakers. As can bee seen on Figure \ref{corpus_sex}, the average F0 value for women population is considerably higher, which is an expected but nevertheless empirically proven result thanks to the availability of structural, speaker and time annotation in SpokesBiz.

\section{Acknowledgements}

The development of SpokesBiz was financed under the 2014-2020 Smart Growth Operational Programme, POIR.04.02.00-00C002/19 for the CLARIN-BIZ project.



\section{References}\label{reference}

\bibliographystyle{lrec}
\bibliography{lrec2020W-xample-kc}


\end{document}